\documentclass[letterpaper]{article} 

\ifdefined\aaaianonymous
    \usepackage[submission]{aaai2026}  
\else
    \usepackage{aaai2026}              
\fi

\usepackage{times}  
\usepackage{helvet}  
\usepackage{courier}  
\usepackage[hyphens]{url}  
\usepackage{graphicx} 
\usepackage{natbib}  
\usepackage{caption} 
\usepackage{algorithm}
\usepackage{algpseudocode}

\usepackage{amssymb}
\usepackage{amsmath}
\usepackage{booktabs}
\usepackage{multirow}

\usepackage{newfloat}
\usepackage{listings}
\DeclareCaptionStyle{ruled}{labelfont=normalfont,labelsep=colon,strut=off} 
\floatstyle{ruled}
\newfloat{listing}{tb}{lst}{}
\floatname{listing}{Listing}

\nocopyright

\ifdefined\aaaianonymous
    \title{Mem4D: Decoupling Static and Dynamic Memory for Dynamic Scene Reconstruction}
\else
    \title{Mem4D: Decoupling Static and Dynamic Memory for Dynamic Scene Reconstruction}
\fi

\author{
Xudong Cai\textsuperscript{\rm 1}, 
Shuo Wang\textsuperscript{\rm 1}, 
Peng Wang\textsuperscript{\rm 1}, 
Yongcai Wang\textsuperscript{\rm 1}\thanks{Corresponding authors}, 
Zhaoxin Fan\textsuperscript{\rm 2}\footnotemark[1], 
Wanting Li\textsuperscript{\rm 1}, 
Tianbao Zhang\textsuperscript{\rm 3}, 
Jianrong Tao\textsuperscript{\rm 4}, 
Yeying Jin\textsuperscript{\rm 5}, 
Deying Li\textsuperscript{\rm 1}
}
\affiliations{
    \textsuperscript{\rm 1}Renmin University of China,
     \textsuperscript{\rm 2}Beihang University,
     \textsuperscript{\rm 3}Shanghai Jiao Tong University, \\
     \textsuperscript{\rm 4}Zhejiang University,
     \textsuperscript{\rm 5}National University of Singapore

}

\begin{document}

\maketitle

\begin{abstract}
Reconstructing dense geometry for dynamic scenes from a monocular video is a critical yet challenging task.
    Recent memory-based  methods enable efficient online reconstruction, but they fundamentally suffer from a Memory Demand Dilemma:
    The memory representation faces an inherent conflict between the long-term stability required for static structures and the rapid, high-fidelity detail retention needed for dynamic motion.
    This conflict forces existing methods into a compromise, leading to either geometric drift in static structures or blurred, inaccurate reconstructions of dynamic objects.
    To address this dilemma, we propose Mem4D, a novel framework that decouples the modeling of static geometry and dynamic motion. Guided by this insight, we design a dual-memory architecture: 
    1) The Transient Dynamics Memory (TDM) focuses on capturing high-frequency motion details from recent frames, enabling accurate and fine-grained modeling of dynamic content;
    2) The Persistent Structure Memory (PSM) compresses and preserves long-term spatial information, ensuring global consistency and drift-free reconstruction for static elements.
    By alternating queries to these specialized memories, Mem4D simultaneously maintains static geometry with global consistency and reconstructs dynamic elements with high fidelity.
    Experiments on challenging benchmarks demonstrate that our method achieves state-of-the-art or competitive performance while maintaining high efficiency. Codes will be publicly available.
\end{abstract}

\ifdefined\aaaianonymous
\else
\fi

\section{Introduction}
The reconstruction of dense dynamic scene geometry from monocular video plays a key role in various real-world applications, such as autonomous driving~\cite{Drivinggaussian}, virtual reality~\cite{Vr-gs}, and robotic navigation~\cite{Vr-robo}.
This task is challenging due to the inherent complexity of dynamic scenes, where camera ego-motion is entangled with the independent movement of dynamic elements.
Traditional SfM~\cite{colmap, agarwal2011building} and SLAM~\cite{Orb-slam3, Vins-mono} methods struggle in dynamic scenes, which violate their core static-world assumption.
Classical approaches handle dynamic scenes by segmenting and filtering out moving objects or modeling their motion independently~\cite{DS-SLAM, DynaSLAM}, but this multi-stage process is complex and prone to error accumulation.

To overcome these limitations, recent research has shifted towards end-to-end learning, producing powerful pointmap regression frameworks.
For instance, DUSt3R~\cite{DUSt3R} directly predicts pair-wise 3D pointmaps from image pairs and conducts a global alignment to assemble a global point cloud, demonstrating impressive performance for static scene reconstruction. Subsequent works like MonST3R~\cite{MonST3R}, D2USt3R~\cite{D2USt3R}, and St4RTrack~\cite{St4RTrack} adapt this architecture for dynamic scenes, achieving notable gains by fine-tuning on dynamic data, adding separate supervision for static and dynamic elements or jointly predicting pointmaps with 3D tracks.
However, their reliance on the costly global alignment remains a bottleneck. To bypass this limitation, some methods process all images in a single pass~\cite{VGGT, Fast3R} at the cost of significant computational overhead and offline processing.
Memory-based methods~\cite{CUT3R, Spann3R} offer a more practical alternative, enabling online reconstruction of long video by incrementally updating a persistent feature memory.
While efficient, their unified memory design struggles to effectively model dynamic scenes, leading to a trade-off between geometric stability and motion fidelity.

\begin{figure}[t]
    \centering
    \includegraphics[width=0.9\columnwidth]{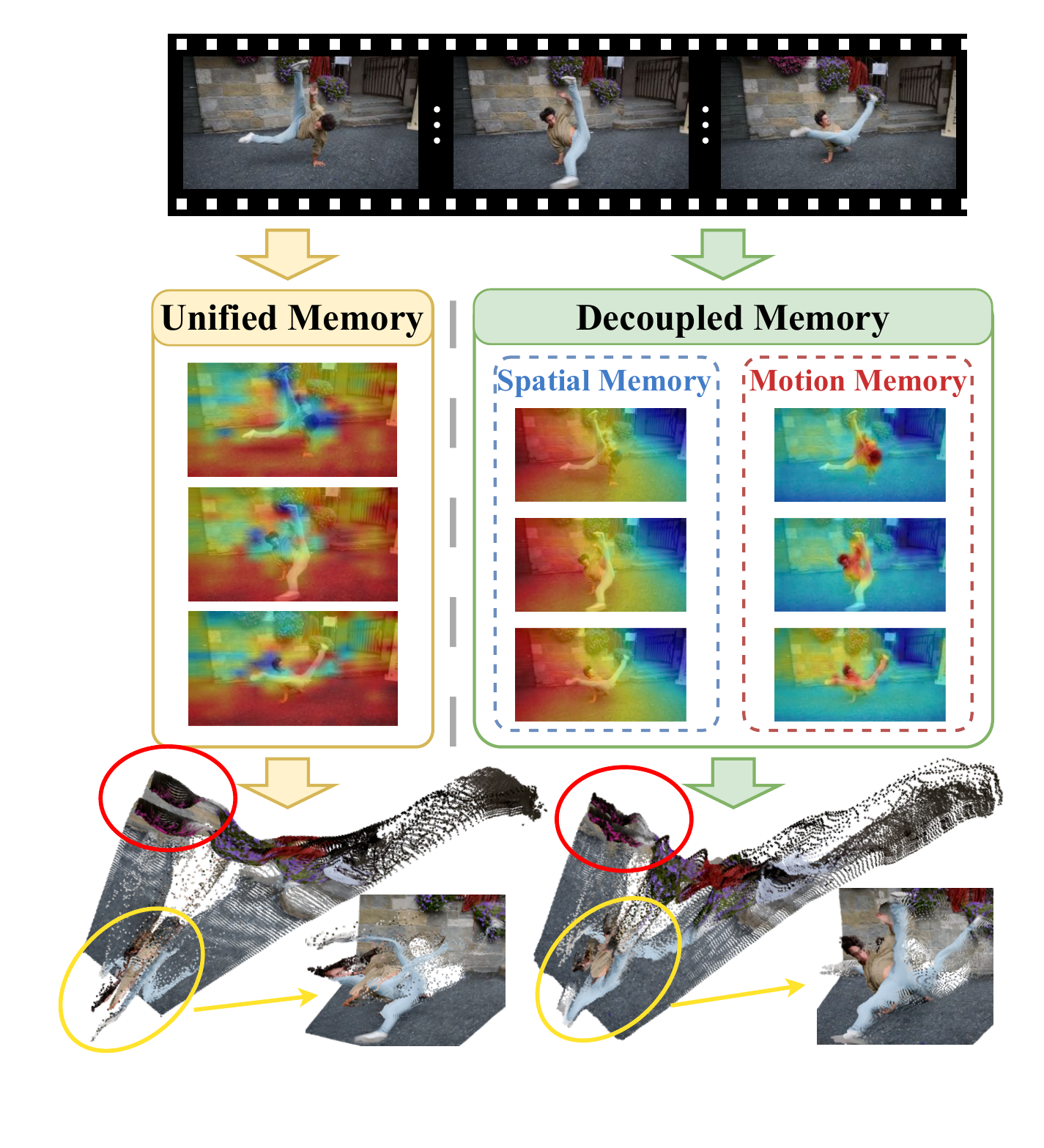}
    \caption{Illustration of the Memory Demand Dilemma.
        We visualize the memory feature maps of several frames. A Unified Memory (left) struggles to represent both static and dynamic elements, resulting in entangled memory features. This leads to geometric drift on the wall (red circle) and severe motion blur on the dancer (yellow circle). In contrast, our Decoupled Memory (right) resolves this conflict by explicitly decoupling the memory into a Spatial Memory for static structures and a Motion Memory for the dancer's movements.
        Our method yields a reconstruction that is both geometrically stable and rich in sharp dynamic details.}
    \label{fig:intro_problem}
\end{figure}

We find the core issue causing this trade-off is a ``Memory Demand Dilemma'': a unified memory is inherent inability to simultaneously provide the long-term consistency required for static structures and the high-fidelity plasticity needed for dynamic motion.
As shown in Figure~\ref{fig:intro_problem}, we can consider the scenario of a person performing a street dance in front of a wall. To preserve the geometric stability of the wall, the memory must enforce long-term consistency, inevitably smearing the dancer's swift movements into a blurry mess. In contrast, to capture these fine movements with high fidelity, the memory must be highly responsive to transient details, ultimately causing the static wall to distort and drift. Consequently, methods relying on a unified memory face an inescapable compromise between geometric stability and motion fidelity.

Inspired by this analysis, we introduce Mem4D, a novel framework for online dynamic reconstruction that resolves the Memory Demand Dilemma. The basic idea is illustrated in Figure~\ref{fig:intro_problem}.
Our key insight is that the static geometry and dynamic motion have different properties and should be modeled separately. Specifically, we introduce a dual-memory architecture that consists of:
(1) a Transient Dynamics Memory (TDM) that captures high-frequency fine-grained motion details by computing 4D cost volumes between current and recent frames, ensuring short-term fidelity.
(2) a Persistent Structure Memory (PSM) that maintains a feature bank of the scene's geometry over time, and employs a spatio-temporal attention mechanism to encode and compress the low-frequency static geometry, ensuring long-term stability.
(3) a Temporal Context Aggregator (TCA) that aggregates rich spatio-temporal context from a local history into the current frame, providing a motion-aware input for our dual memories.

In contrast to previous methods, our decoupled design eliminates the memory conflict inherent in unified memories. This allows Mem4D to synthesize a superior reconstruction by fusing the global stability from the PSM with the sharp motion details from the TDM. Experiments on challenging benchmarks show that Mem4D achieves competitive or state-of-the-art performance in dynamic scene reconstruction, while keeping high efficiency.

The main contributions of our work are as follows:
\begin{itemize}
    \item We introduce Mem4D, a novel online dual-memory framework for dynamic scene reconstruction by decoupling static geometry and dynamic motion.
    \item We introduce the Transient Dynamics Memory to preserve high-fidelity motion details and the Persistent Structure Memory with spatio-temporal compression to ensure long-term geometric stability.
    \item Our method achieves state-of-the-art or competitive performance on challenging benchmarks, delivering superior accuracy for static structures while preserving fine-grained dynamic motion.
\end{itemize}

\section{Related Work}

\subsection{Static 3D Reconstruction}
Classical Static 3D reconstruction primarily relies on Structure from Motion (SfM)~\cite{agarwal2011building, colmap} and Simultaneous Localization and Mapping (SLAM)~\cite{Orb-slam3,Vins-mono}. These methods are largely rooted in multi-stage geometric pipelines such as keypoints matching~\cite{Harris, SIFT} and bundle adjustment~\cite{BA}. Despite their widespread success, they remain susceptible to failure in degenerate scenarios (e.g., low-texture regions or views with minimal overlap).

Deep learning has progressively revolutionized the field. Initially, this involved replacing handcrafted modules with learnable ones, such as feature matching~\cite{Superglue, LoFTR}. While improving specific stages, the overall process remained fragmented and susceptible to error propagation.
This motivated a paradigm shift towards end-to-end frameworks. While early works~\cite{DeepSfM, VGGSfM} introduced differentiable SfM pipelines, they often struggled with generalization and global consistency. A significant breakthrough came with pointmap regression models like DUSt3R~\cite{DUSt3R}. They directly map images to 3D pointmaps using transformer~\cite{Transformer}, achieving impressive results on static scenes. However, their pairwise design requires a costly global alignment to assemble multiple pointmaps, limiting their scalability.

To address this, some works process all images in a single pass~\cite{VGGT, Fast3R, Mvdust3r, Light3R, Dens3R, PI3} at the cost of significant computational overhead and offline processing. In contrast, online methods have emerged as a more practical paradigm. Spann3R~\cite{Spann3R} pioneered this direction by introducing an external memory bank that is incrementally updated as new frames arrive. Subsequent works~\cite{CUT3R, SLAM3R, Point3R, MUSt3R} focused on developing more elaborate designs for the unified memory architecture. While effective for static scenes, these methods struggle with dynamic scenes. We address this limitation by introducing a dual-memory architecture that decouples the modeling of static geometry and dynamic motion, achieving superior performance.

\subsection{Dynamic Scene Reconstruction}
Traditional approaches for dynamic scenes often rely on complex multi-stage pipelines involving explicit object segmentation and tracking~\cite{DS-SLAM, DynaSLAM, MaskFusion} or per-scene optimization with depth priors~\cite{Robust-CVD, CasualSAM, MegaSaM}. While capable, their complexity makes them prone to error accumulation.

To address this issue, the recent paradigm of end-to-end pointmap regression has been adapted for dynamic scenes in various ways, such as introducing new training strategies~\cite{MonST3R, Stereo4D, D2USt3R}, geometry and generative priors~\cite{Sora3R, Geo4D, Align3R}, attention adaptation during inference~\cite{Easi3R} or joint tracking formulations~\cite{St4RTrack, DPM, POMATO}. However, these methods still require a costly global alignment, limiting their practicality for long videos.
To bypass this limitation, recent works have begun to incorporate memory design: Driv3R~\cite{Driv3R} extends spatial memory to the temporal dimension, CUT3R~\cite{CUT3R} uses recurrent states to encode scene history, and MUSt3R~\cite{MUSt3R} leverages a multi-layer memory mechanism to reduce the computational complexity. Concurrent work Point3R~\cite{Point3R} proposes explicit 3D pointer-based memories and StreamVGGT~\cite{StreamVGGT} leverages implicit key-value caches from causal transformers.

Despite these varied and elaborate designs, a common limitation persists: they all rely on a single, unified memory to encode both static geometry and dynamic motions. This unified memory inevitably faces a fundamental conflict between preserving long-term geometric stability and capturing high-fidelity motion. We propose a novel dual-memory architecture to resolve this conflict by decoupling the modeling of static geometry and dynamic motion, resulting in superior reconstruction performance.

\subsection{Memory Bank in Computer Vision}
Memory banks are widely studied in computer vision for their ability to store and retrieve historical information. This ability makes them suitable for sequential tasks like video object segmentation~\cite{STM,Xmem}, video recognition~\cite{Memvit}, optical flow estimation~\cite{Memflow} and video generation~\cite{FramePack}. For instance, FramePack~\cite{FramePack} introduces a memory structure that progressively compresses input frames based on their importance to improve video generation fidelity. This concept shares a similar spirit with our Persistent Structure Memory, which compresses the historical memory features based on their temporal distance for long-term stability. However, FramePack compresses the input frames to improve efficiency, while our PSM compresses past pointmaps for geometry stability.

\section{Method}

\begin{figure*}[htbp]
    \includegraphics[width=\textwidth]{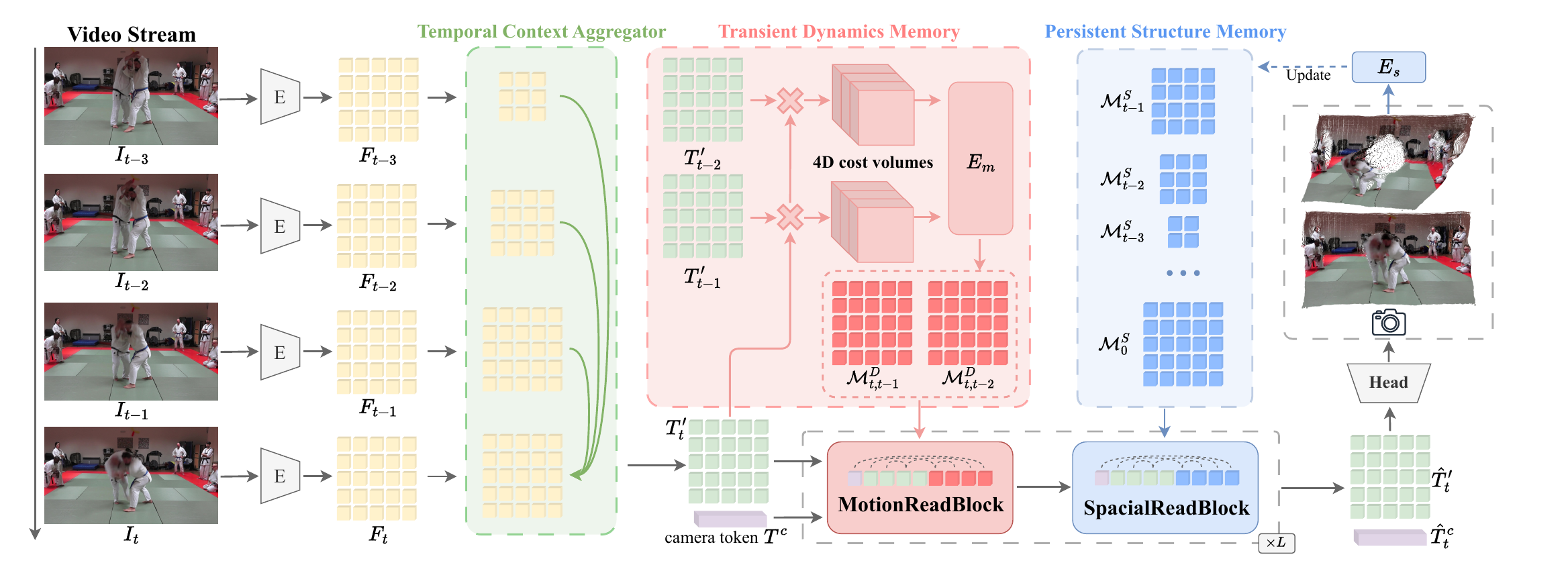}
    \caption{Framework of Mem4D. For each incoming frame $I_t$, the TCA first enriches the ViT-encoded $F_t$ by aggregating features from a local history to create a motion-aware representation $T^{\prime}_t$. Then, $T^{\prime}_t$ and a learnable camera token $T^c$ are iteratively refined by interleaved readouts from the TDM and PSM via self-attention. The TDM is computed on-the-fly from 4D correlation volumes with recent feature maps, capturing fine-grained motion cues. Concurrently, the PSM maintains a long-term FIFO feature bank of the scene's geometry to ensure global consistency. All features but the initial one $\mathcal{M}_0^S$ are compressed based on their temporal distance to $t$. Finally, the refined tokens $\hat{T}_t^c$ and $\hat{T}_t^{\prime}$ are passed to prediction heads to output the global and self-view pointmaps, along with camera parameters for the current frame. The global pointmap is encoded to update the PSM.
    }
    \label{fig:method_overview}
\end{figure*}

Given a monocular video stream $\mathcal{I}=\left\{I_1, I_2, \ldots, I_T\right\}$ of a dynamic scene, our goal is to reconstruct a sequence of dense 3D pointmaps $\mathcal{X}=\left\{X_1, X_2, \ldots, X_T\right\}$ and camera parameters $\mathcal{P}=\left\{\left(K_1, T_1\right),\left(K_2, T_2\right), \ldots,\left(K_T, T_T\right)\right\}$ within a consistent global coordinate frame and $K_i$ is the camera intrinsic.
Figure~\ref{fig:method_overview} illustrates the overview of Mem4D.
Our key contribution is a dual-memory architecture that models static geometry and dynamic motion separately, allowing the model to draw upon distinct information sources for geometric stability and motion fidelity.

In the following sections, we first introduce the Temporal Context Aggregator, which enriches each input frame with recent motion context. Next, we elaborate on the core of our framework: the Transient Dynamics Memory for capturing high-fidelity dynamics, and the Persistent Structure Memory for maintaining long-term structural stability. Finally, we present our training strategy and objectives.

\subsection{Temporal Context Aggregator (TCA)}
\label{Sec: TCA}
A single video frame is often insufficient to resolve motion ambiguities. Therefore, to create a more rich input for the subsequent memory fusion, we first enrich the current frame's features with its temporal neighborhood using TCA.

For each incoming frame $I_t$, we first extract a feature map $F_t \in \mathbb{R}^{H \times W \times C}$ using a ViT~\cite{ViT} encoder. The TCA enriches  $F_t$ by aggregating context from a local window of $k_t$ past frames $\left\{F_{t-j}\right\}_{j=1}^{k_t}$.
To efficiently summarize this history, we employ a distance-aware temporal compression scheme. Each past feature map $F_{t-j}$ is spatially downsampled using a 2D convolution $\mathrm{Conv}_{s_j}$,  where the stride $s_j = \phi(j)$ is a function of its temporal distance $j$:
\begin{equation}
    s_j = \phi(j) = \begin{cases} 1, & \text{if } 0 \le j < 2 \\ 2, & \text{if } 2 \le j < 4 \\ 4, & \text{if } j \ge 4 \end{cases}
\end{equation}
The compressed feature map $\tilde{F}_{t-j}$ is thus given by:
\begin{equation}
    \tilde{F}_{t-j} = \text{Conv}_{\phi(j)}(F_{t-j}) \quad \forall j \in \{1, \dots, k_t\}
\end{equation}
This scheme progressively summarizes older features while preserving the high-fidelity detail of recent ones.
Then, the current image tokens $F_t$ are concatenated with the compressed past tokens $\left\{\tilde{F}_{t-j}\right\}_{j=1}^{k_t}$ and processed by four self-attention layers. To accurately capture the spatio-temporal relationships that are crucial for motion, we incorporate 3D Rotary Position Embeddings (3DRoPE)~\cite{RoPE} to jointly encode each token's position.
The self-attention allows the current frame tokens to draw relevant context from the aggregated history, producing the enriched tokens $T_t^{\prime}$.

\subsection{Decoupled Static and Dynamic Memory}
\label{Sec: Memory}
The core of Mem4D is a dual-memory architecture that models static geometry and dynamic motion separately, enabling drift-free and motion-fidelity reconstruction. We first describe how each memory is constructed, then describe the memory readout process.

\paragraph{Transient Dynamics Memory (TDM)}
Motion consists of high-frequency signals whose relevance is local in time.
To capture fine-grained motion details without loss, we introduce the Transient Dynamics Memory. The TDM  stores $k_d$ motion-aware feature maps, denoted as  $\mathcal{M}^D \in \mathbb{R}^{k_d \times H \times W \times C_m}$. Each feature map is derived from a 4D correlation volume $C_{t,t-j}$. $C_{t,t-j}$ is computed by taking the dot product between the current features $T_t^{\prime}$ and a past feature $T_{t-j}^{\prime}$ from the set of past frames $\left\{T^{\prime}_{t-j}\right\}_{j=1}^{k_d}$:
\begin{equation}
    C_{t,t-j} = T_t^{\prime} \times (T_{t-j}^{\prime})^T \in \mathbb{R}^{H \times W \times H \times W}.
\end{equation}
Following RAFT~\cite{RAFT}, we build a multi-scale correlation pyramid $\left\{C^{l}_{t,t-j}\right\}_{l=1}^{L}$ by pooling the last two dimensions of the correlation volume at different scales. This pyramid, which captures both large and small displacements, is then concatenated along the channel dimension and projected by an MLP to produce a motion feature map:
\begin{equation}
    M_{t,t-j} = \text{MLP}\left(C^{1}_{t,t-j}\oplus	 C^{2}_{t,t-j}, \ldots, C^{L}_{t,t-j}\right).
\end{equation}
where $\oplus$ denotes the concatenation operation. A motion feature encoder $E_m$ consisting of four self-attention layers with 3DRoPE, further refines $M_{t,t-j}$ to yield the final motion memory feature $\mathcal{M}^D_{t,t-j} = E_m\left(M_{t,t-j}\right) \in \mathbb{R}^{H \times W \times C_m}$.
Unlike a persistent memory, the TDM is computed on-the-fly at each timestep $t$, storing motion-aware features relevant only to the current context.

\begin{table*}[htbp]
    \centering
    \small
    \renewcommand{\arraystretch}{0.9}
    \renewcommand{\tabcolsep}{3pt}
    \begin{tabular}
        {@{}ccccccccc@{}}
        \toprule
                           &                 &               & \multicolumn{2}{c}{\textbf{Sintel}} & \multicolumn{2}{c}{\textbf{BONN}}  & \multicolumn{2}{c}{\textbf{KITTI}}                                                                                                     \\
        \cmidrule(lr){4-5} \cmidrule(lr){6-7} \cmidrule(lr){8-9}
        \textbf{Alignment} & \textbf{Method} & \textbf{Type} & {Abs Rel $\downarrow$}              & {$\delta$\textless $1.25\uparrow$} & {Abs Rel $\downarrow$}             & {$\delta$\textless $1.25\uparrow$} & {Abs Rel $\downarrow$} & {$\delta$ \textless $1.25\uparrow$} \\
        \midrule
        \multirow{6}{*}{\begin{minipage}[c]{1.5cm}Per-Scene\end{minipage}}
                           & DUSt3R-GA       & PW            & 0.656                               & \underline{45.2}                   & \underline{0.155}                  & \underline{83.3}                   & \textbf{0.144}         & \underline{81.3}                    \\
                           & MASt3R-GA       & PW            & 0.641                               & 43.9                               & 0.252                              & {70.1}                             & 0.183                  & 74.5                                \\
                           & MonST3R-GA      & PW            & \textbf{0.378}                      & \textbf{55.8}                      & \textbf{0.067}                     & \textbf{96.3}                      & 0.168                  & 74.4                                \\

                           & Fast3R          & FF            & \underline{0.628}                   & 43.8                               & 0.193                              & 77.3                               & \underline{0.152}      & \textbf{84.1}                       \\
        \cmidrule(l){2-9}
                           & Spann3R         & OL            & 0.622                               & 42.6                               & 0.144                              & {81.3}                             & 0.198                  & 73.7                                \\
                           & CUT3R           & OL            & \textbf{0.421}                      & \textbf{47.9}                      & \underline{0.078}                  & \underline{93.7}                   & \textbf{0.118}         & \textbf{88.1}                       \\
                           & \textbf{Ours}   & OL            & \underline{0.520}                   & \underline{43.1}                   & \textbf{0.072}                     & \textbf{95.7}                      & \underline{0.140}      & \underline{82.0}                    \\
        \midrule
        \multirow{3}{*}{\begin{minipage}[c]{1.5cm}Metric\end{minipage}}
                           & MASt3R-GA       & PW            & \underline{1.022}                   & 14.3                               & 0.272                              & 70.6                               & 0.467                  & 15.2                                \\
                           & CUT3R           & OL            & 1.029                               & \textbf{23.8}                      & \underline{0.103}                  & \underline{88.5}                   & \underline{0.122}      & \textbf{85.5}                       \\
                           & \textbf{Ours}   & OL            & \textbf{0.846}                      & \underline{22.3}                   & \textbf{0.086}                     & \textbf{95.7}                      & \textbf{0.117}         & \underline{79.5}                    \\
        \bottomrule
    \end{tabular}
    \caption{Video Depth Evaluation.
        We compare scale-invariant depth (Per-Scene alignment) and metric depth (no alignment) results on Sintel, Bonn, and KITTI datasets. PW denotes pair-wise methods, OL denotes online methods, FF denotes feed-forward methods, and GA denotes global alignment. The best results are in \textbf{bold} and the second best are \underline{underlined}.}
    \label{tab:videodepth}
\end{table*}

\paragraph{Persistent Structure Memory (PSM)}
To ensure global consistency and prevent long-term drift, the Persistent Structure Memory maintains a feature bank of the scene's structure over time, denoted as $\mathcal{M}^S \in \mathbb{R}^{k_s \times H \times W \times C_s}$.
At each timestep $t$, the PSM is updated by appending a new feature map $\mathcal{M}^S_t$. $\mathcal{M}^S_t$ is generated using a lightweight encoder $E_s$ which is built primarily of four self-attention blocks. $E_s$ processes the final predicted global pointmap $\hat{X}^{global}_t$ and transforms it into a compact feature representation that encapsulates the scene's essential geometric structure: $\mathcal{M}^S_t = E_s\left(\hat{X}^{global}_t\right) \in \mathbb{R}^{H \times W \times C_s}$.
This memory is managed as a First-In-First-Out (FIFO) queue of size $k_s$. Note that the first memory frame $\mathcal{M}^S_0$ is never removed to serve as a stable global anchor against drift.

\paragraph{Memory Readout}
The final reconstruction is produced by an iterative fusion decoder. It progressively refines the current frame's tokens $T_t^{\prime}$ and a learnable camera token $T_t^c$ over $L$ stages of interleaved readouts from the TDM and PSM, as detailed in Algorithm~\ref{algo:readout}.
Each block first employs a MotionReadBlock to refine the current tokens by attending to the TDM to fuse high-frequency motion features. Subsequently, a SpatialReadBlock updates the tokens by querying the PSM, grounding the prediction in a stable, long-term geometric context. Both blocks are implemented as self-attention layers enhanced with 3DRoPE to effectively encode spatio-temporal relationships. The alternating design enables the model to first resolve fine-grained dynamics and then anchor them within a globally consistent framework.

Prior to the readout stage, the two memories are handled differently. The TDM is directly used for readout to retain maximum detail as it captures high-frequency, transient motion information. In contrast, the PSM undergoes a temporally-aware compression to filter noise and reduce redundancy, as it stores low-frequency, long-term geometric information. We apply 3D convolutions with varying kernel sizes $s_d$ to the memory frame $\mathcal{M}^S_j$ based on its temporal distance $d = t-j$ to the current frame $t$:
\begin{equation}
    \label{eq:compression}
    s_d = \begin{cases}
        (4, 8, 8), & \text{if } d \ge 6     \\
        (2, 4, 4), & \text{if } 4 \le d < 6 \\
        (1, 2, 2), & \text{if } 2 \le d < 4 \\
        (1, 1, 1), & \text{otherwise}
    \end{cases}
\end{equation}
This policy enables the model to effectively keep long-term structural cues and preserve high-fidelity detail for recent ones. Crucially, the initial frame $\mathcal{M}^S_0$ is never compressed, serving as a stable, high-resolution anchor to mitigate drift.
\begin{algorithm}
    \caption{Iterative Memory Readout}
    \begin{algorithmic}[1]
        \State \textbf{Input:} Current frame tokens $T_t^{\prime(0)}$, Pose Token $T_t^{c(0)}$, TDM $\mathcal{M}^{D(0)}$ and compressed PSM $\mathcal{M}^{S(0)}$.
        \For{$l=1$ \textbf{to} $L$}
        \State $\tilde{T}_t^{c(l)}$, $\tilde{T}_t^{\prime(l)}$, $\mathcal{M}^{D(l)}_{t}$ $\leftarrow$\text{MotionReadBlock}($T_t^{c(l-1)}$, $T_t^{\prime(l-1)}$, $\mathcal{M}^{D(l-1)}$)
        \State $T_t^{c(l)}$, $T_t^{\prime(l)}$, $\mathcal{M}^{S(l)}_{t}$ $\leftarrow$\text{SpacialReadBlock}($\tilde{T}_t^{c(l)}$, $\tilde{T}_t^{\prime(l)}$, $\mathcal{M}^{S(l-1)}$)
        \EndFor
        \State \textbf{return} $T_t^{c(L)}$, $T_t^{\prime(L)}$
    \end{algorithmic}
    \label{algo:readout}
\end{algorithm}
After $L$ iterations, the current frame tokens $\hat{T}_t^{\prime} = T_t^{\prime(L)}$ are processed by two DPT heasd~\cite{DPT} to predict global and self-view coordinate pointmaps $\hat{X}^{global}_t$ and $\hat{X}^{self}_t$, along with their corresponding confidence maps $C^{global}_t$ and $C^{self}_t$. The camera token $\hat{T}_t^c = T_t^{c(L)}$ is fed to an MLP head to regress the intrinsics (FoV) and camera pose (quaternion and translation vector):

 \begin{equation} \label{eq:prediction_heads}
     \begin{aligned}
         \hat{X}^{global}_t, C^{global}_t & = \text{Head}_{global}\left(\hat{T}_t^{\prime}\right), \\
         \hat{X}^{self}_t, C^{self}_t     & = \text{Head}_{self}\left(\hat{T}_t^{\prime}\right),   \\
         \hat{K}_t, \hat{T}_t             & = \text{Head}_{camera}\left(\hat{T}_t^c\right).
     \end{aligned}
 \end{equation}

\begin{figure*}[htbp]
    \centering
    \includegraphics[width=0.7\linewidth]{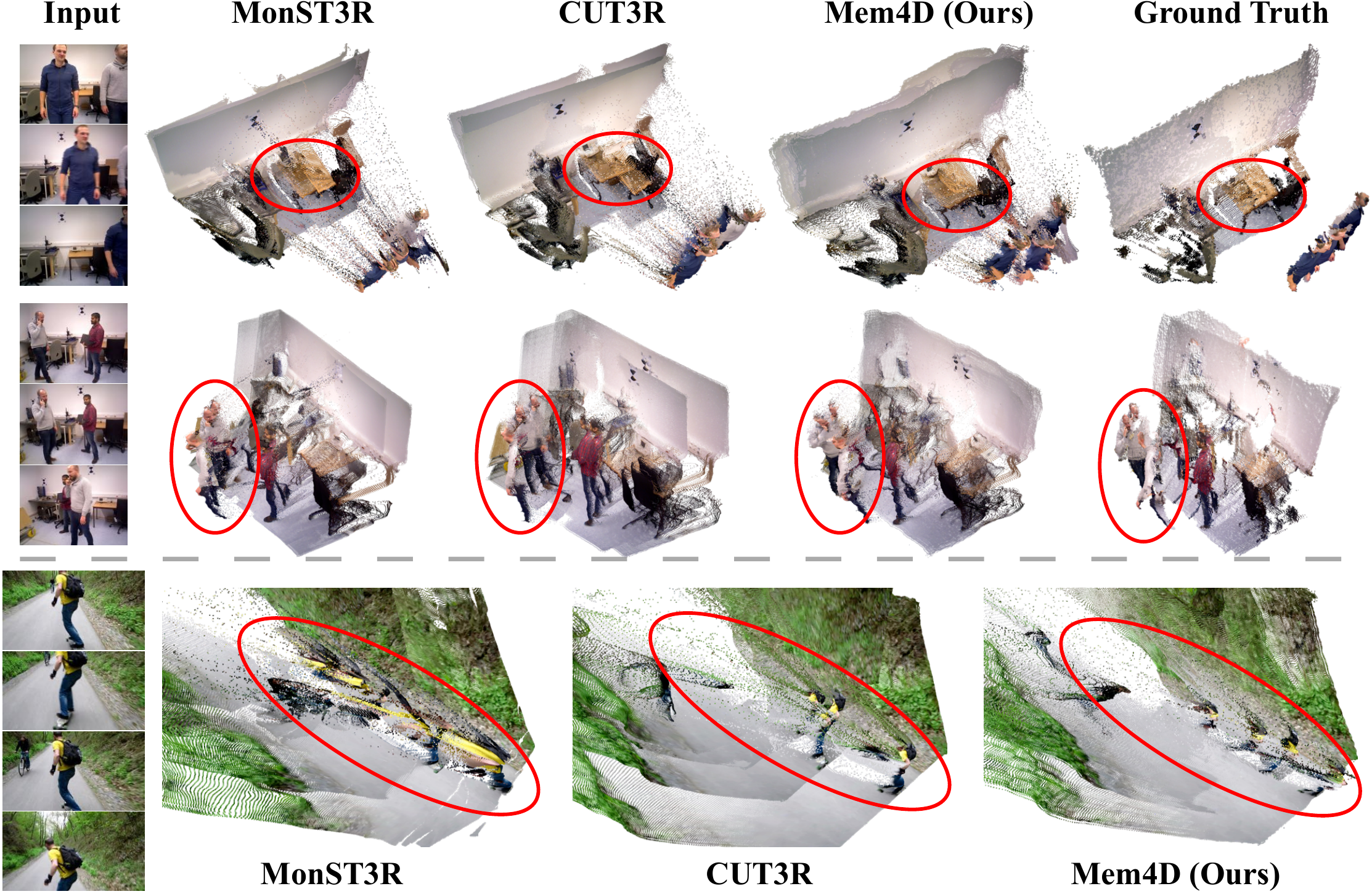}
    \caption{Qualitative results of dynamic reconstruction on Bonn (top) and DAVIS (down) dataset.  Compared to MonST3R and CUT3R, our Mem4D achieves the best qualitative results. }
    \label{fig:qualitative_depth}
\end{figure*}

\subsection{Training Objective and Strategy}
\label{Sec: Training}
Our model is trained end-to-end on sequences of $N$ images. The total loss function is a weighted sum of losses for pointmap reconstruction and camera parameter estimation.
\paragraph{3D regression loss.} We denote the predicted pointmaps as $\mathcal{X} = \left\{\hat{X}^{global}_1, \ldots, \hat{X}^{global}_N, \hat{X}^{self}_1, \ldots, \hat{X}^{self}_N\right\}$ and their corresponding confidence scores as $\mathcal{C}$.
We adopt the confidence-aware regression loss following MASt3R~\cite{MASt3R}.
\begin{equation}
    \mathcal{L}_{\mathrm{conf}}=\sum_{(\hat{\boldsymbol{x}}, c) \in(\hat{\mathcal{X}}, \mathcal{C})}\left(c \cdot\left\|\frac{\hat{\boldsymbol{x}}}{s}-\frac{\boldsymbol{x}}{s}\right\|_2-\alpha \log c\right)
\end{equation}
The scale normalization factor $s$ is from the groundtruth, enabling the model to learn metric-scale pointmaps.

\paragraph{Camera Loss.}
The camera parameter is supervised with two components. First, we apply a L2 loss on the predicted absolute pose (quaternion $\hat{\boldsymbol{q}}_t$ and translation $\hat{\boldsymbol{\tau}}_t$) and camera intrinsic $\hat{K}$ for each frame against their ground-truth values:
\begin{equation}
    \small
    \mathcal{L}_{\text {abspose}}=\sum_{t=1}^N\left(\left\|\hat{\boldsymbol{q}}_t-\boldsymbol{q}_t\right\|_2+\left\|\frac{\hat{\boldsymbol{\tau}}_t}{s}-\frac{\boldsymbol{\tau}_t}{s}\right\|_2 + \beta\left\| K_t -\hat{K}_t \right\| \right)
\end{equation}
To improve temporal consistency, we introduce an L2 loss on the relative pose between consecutive frames. We compute the predicted relative pose $\hat{T}_{t, t-1} = \hat{T}_t \times \hat{T}_{t-1}^{-1}$  and supervise it using the ground-truth relative:
\begin{equation}
    \small
    \mathcal{L}_{\text{relpose}} = \sum_{t=2}^{N} \left( \| \hat{\boldsymbol{q}}_{t,t-1} - \boldsymbol{q}_{t,t-1} \|_2 +\left\|\frac{\hat{\boldsymbol{\tau}}_{t,t-1}}{s}-\frac{\boldsymbol{\tau}_{t,t-1}}{s}\right\|_2 \right)
\end{equation}
The final training objective is the weighted sum of these components:
\begin{equation}
    \mathcal{L} = \lambda_1 \mathcal{L}_{\text{conf}} + \lambda_2 \mathcal{L}_{\text{abspose}} + \lambda_3 \mathcal{L}_{\text{relpose}}
\end{equation}

\begin{table*}[t]
    \centering
    \small
    \renewcommand{\arraystretch}{0.9}
    \renewcommand{\tabcolsep}{3pt}
    \begin{tabular}{ccccccccccc c}
        \toprule
                          &               & \multicolumn{3}{c}{\textbf{Stereo4D}} & \multicolumn{3}{c}{\textbf{Sintel}} & \multicolumn{3}{c}{\textbf{TUM-dynamics}} & \multirow{2}{*}{\textbf{FPS $\uparrow$}}                                                                                                                                     \\
        \cmidrule(lr){3-5} \cmidrule(lr){6-8} \cmidrule(lr){9-11}
        {\textbf{Method}} & \textbf{Type} & {ATE $\downarrow$}                    & {RPE trans $\downarrow$}            & {RPE rot $\downarrow$}                    & {ATE $\downarrow$}                       & {RPE trans $\downarrow$} & {RPE rot $\downarrow$} & {ATE $\downarrow$} & {RPE trans $\downarrow$} & {RPE rot $\downarrow$} &      \\
        \midrule
        DUSt3R-GA         & PW            & {0.931}                               & 0.896                               & 4.541                                     & 0.417                                    & 0.250                    & 5.796                  & \underline{0.083}  & \underline{0.017}        & 3.567                  & \textbf{0.33} \\
        MASt3R-GA         & PW            & {\bf{0.382}}                          & {\bf{0.1951}}                       & {\bf{0.687}}                              & \underline{0.185}                        & \underline{0.060}        & \underline{1.496}      & {\bf{0.038}}       & {\bf{0.012}}             & {\bf{0.448}}           & \underline{0.15} \\
        MonST3R-GA        & PW            & {\underline{0.414}}                   & {\underline{0.631}}                 & {\underline{1.052}}                       & \bf{{0.111}}                             & \bf{0.044}               & \bf{0.869}             & {{0.098}}          & {{0.019}}                & {\underline{0.935}}    & 0.11 \\
        \midrule
        Spann3R           & OL            & 0.653                                 & 0.321                               & 1.353                                     & {{0.329}}                                & 0.110                    & 4.471                  & \underline{0.056}  & 0.021                    & 0.591                  & 9.07 \\
        CUT3R             & OL            & \underline{0.506}                     & \underline{0.244}                   & \underline{0.730}                         & \textbf{0.213}                           & \textbf{0.066}           & \textbf{0.621}         & \textbf{0.046}     & \textbf{0.015}           & \textbf{0.473}         &   \textbf{17.91}   \\
        \textbf{Ours}     & OL            & \textbf{0.495}                        & \textbf{0.229}                      & \textbf{0.641}                            & \underline{0.263}                        & \underline{0.091}        & \underline{0.812}      & 0.061              & \underline{0.020}        & \underline{0.517}      &    \underline{16.12}  \\
        \bottomrule
    \end{tabular}
    \caption{Camera Pose Estimation Evaluation on ScanNet, Sintel, and TUM-dynamics datasets. PW denotes pair-wise method, OL denotes online method and GA denotes global alignment. The best results are in \textbf{bold} and the second best are \underline{underlined}. We report FPS on  Stereo4D  at $512 \times 384$  resolution for all methods, except Spann3R which only supports $224 \times 224$ images. }
    \label{tab:pose}
\end{table*}

\paragraph{Curriculum Training}
We train Mem4D on a diverse mix of 10 datasets, including both synthetic and real-world data such as ARKitScenes~\cite{Arkitscenes} and Spring~\cite{Spring}. See the supplementary material for more details. The image resolution is set to a maximum of 512 pixels on the longer side.
The training is divided into two stages. We first train the model on fixed-length sequences by sampling 11 frames per video sequence. In the second stage, we train the model on longer sequences by randomly sampling 11 to 48 frames from each video sequence, improving the model's ability to handle longer videos.

\paragraph{Implementation Details}
The image encoder is a ViT-Large~\cite{ViT} initialized from CUT3R~\cite{CUT3R} weights and is frozen during training. The memory feature of TDM and PSM is set to 768 channels. The window size $k_t$ of TCA is 5 and the memory sizes $k_d$ and $k_s$ are set to 2 and 100, respectively. We use AdamW~\cite{AdamW} optimizer with an initial learning rate of $1e^{-5}$. Linear warmup and cosine decay are applied. Training is conducted on 8 NVIDIA A100 GPUs with a batch size of 4 per GPU.

\section{Experiment}

We evaluate our method on video depth estimation, camera pose estimation, and 3D scene reconstruction tasks. We compare Mem4D against leading methods which can be broadly categorized into two groups: (1) Offline methods need the entire set of views to form a complete reconstruction, such as DUSt3R~\cite{DUSt3R}, MASt3R~\cite{MASt3R}, MonST3R~\cite{MonST3R} and Fast3R~\cite{Fast3R}; (2) Online methods that process the input frames sequentially, such as Spann3R~\cite{Spann3R} and CUT3R~\cite{CUT3R}.

\subsection{Video Depth Estimation Performance}
Following the protocol of~\cite{MonST3R,CUT3R}, we evaluate depth prediction for long dynamic videos on KITTI~\cite{KITTI}, Sintel~\cite{Sintel} and Bonn~\cite{Bonn} benchmarks. We report the absolute relative error (Abs Rel) and percentage of inlier points $\delta < 1.25$.
We evaluate Mem4D both with alignment (Per-Scene scale) and without alignment (Metric).

As shown in Table~\ref{tab:videodepth}, under Per-Scene alignment, Mem4D performs competitively against CUT3R, notably surpassing it on the challenging Bonn dataset (0.072 vs. 0.078 Abs Rel), and clearly outperforms Spann3R across all benchmarks. While MonST3R achieves state-of-the-art results, its performance relies on a costly global alignment process and requires additional inputs such as optical flow~\cite{RAFT} and semantic segmentation~\cite{SAM2}.
More importantly, in the challenging Metric (no alignment) evaluation, Mem4D establishes a new state of the art, significantly outperforming CUT3R on major benchmarks. On Sintel and Bonn, we improve the absolute relative error by 21.6\% and 19.7\%, respectively. This substantial improvement validates the effectiveness of our design choices. Figure~\ref{fig:qualitative_depth} shows three qualitative examples, demonstrating the superior performance of our method.

\begin{table}[]
    \centering
    \small
    \renewcommand{\arraystretch}{0.9}
    \renewcommand{\tabcolsep}{1.5pt}
    \begin{tabular}{cccccccccccc}
        \toprule
        \multirow{3}{*}{Method} & \multicolumn{5}{c}{7-Scenes}         & \multicolumn{5}{c}{NRGBD}                                                                                                                                                                                                              \\ \cline{2-12}
        \rule{0pt}{8pt}         & \multicolumn{2}{c}{Acc $\downarrow$} &                           & \multicolumn{2}{c}{Comp $\downarrow$} &                   & \multicolumn{2}{c}{Acc $\downarrow$} &  & \multicolumn{2}{c}{Comp $\downarrow$}                                                                \\ \cline{2-3} \cline{5-6} \cline{8-9} \cline{11-12}
        \rule{0pt}{8pt}         & Mean                                 & Med.                      &                                       & Mean              & Med.                                 &  & Mean                                  & Med.              &  & Mean              & Med.              \\
        \midrule
        DUST3R-GA               & \textbf{0.146}                       & \textbf{0.077}            &                                       & \underline{0.181} & \underline{0.067}                    &  & \textbf{0.144}                        & \textbf{0.019}    &  & \textbf{0.154}    & \textbf{0.018}    \\
        MonST3R-GA              & 0.248                                & 0.185                     &                                       & 0.266             & 0.167                                &  & \underline{0.272}                     & \underline{0.114} &  & 0.287             & 0.110             \\
        Fast3R                  & \underline{0.155}                    & \underline{0.104}         &                                       & \textbf{0.125}    & \textbf{0.052}                       &  & 0.366                                 & 0.239             &  & \underline{0.198} & \underline{0.097} \\
        \midrule
        Spann3R                 & 0.298                                & 0.226                     &                                       & 0.205             & 0.112                                &  & 0.416                                 & 0.323             &  & 0.417             & 0.285             \\
        CUT3R                   & \textbf{0.126}                       & \textbf{0.047}            &                                       & \textbf{0.154}    & \textbf{0.031}                       &  & \textbf{0.099}                        & \textbf{0.031}    &  & \textbf{0.076}    & \textbf{0.026}    \\
        Ours                    & \underline{0.185}                    & \underline{0.137}         &                                       & \underline{0.178} & \underline{0.082}                    &  & \underline{0.271}                     & \underline{0.196} &  & \underline{0.212} & \underline{0.113} \\
        \bottomrule
    \end{tabular}
    \caption{3D Reconstruction Evaluation on 7-scenes and NRGBD datasets. The best results are in \textbf{bold} and the second best are \underline{underlined}.}
    \label{tab:reconstruction}

\end{table}

\subsection{Camera Pose Estimation Performance}
Following standard evaluation protocols~\cite{MonST3R,CUT3R}, we evaluate camera pose estimation on the Sintel~\cite{Sintel} and TUM dynamics~\cite{TUM}. We also conduct evaluations on the challenging Stereo4D~\cite{Stereo4D} dataset. For Stereo4D, we report results on 50 sequences randomly sampled from the official test split, sampling keyframes every 5 images. For all evaluations, we report Absolute Translation Error (ATE), Relative Translation Error (RPE trans), and Relative Rotation Error (RPE rot) after Sim(3) alignment with the ground truth.

It is crucial to note that Mem4D is a fully online method without any post-processing, while top methods like MonST3R-GA achieve superior pose accuracy via costly, offline global alignment. As shown in Table~\ref{tab:pose}, Mem4D achieves competitive performance against CUT3R, for instance, delivering a strong accuracy on Stereo4D. We argue this trade-off enables significant efficiency gains: Mem4D runs at 16 FPS, making it suitable for real-time applications.

\subsection{3D Reconstruction Performance}
To demonstrate Mem4D's generalizability on the static scenes, we evaluate our method on the 7-scenes~\cite{7Scenes} and NRGBD~\cite{NRGBD} benchmarks.
Following the evaluation protocols~\cite{CUT3R}, we use sparse inputs: 3-5 frames for 7-scenes and 2-4 frames for NRGBD and report accuracy (Acc) and completion (Comp). Table~\ref{tab:reconstruction} shows Mem4D achieves results comparable to leading online methods, demonstrating the strong generalizability of our method for static scenes.

\subsection{Ablation Study}
We validate our design choices with the ablation study presented in Table~\ref{tab:ablation}. This table evaluates several smaller-scale model variants on the Sintel dataset for video depth and pose accuracy, with each variant corresponding to a specific row. Row one is our full implementation. All models were trained on ARKitScenes~\cite{Arkitscenes} and PointOdyssey~\cite{PointOdyssey} under consistent settings.

\begin{table}[]
    \centering
    \small
    \renewcommand{\arraystretch}{0.85}
    \renewcommand{\tabcolsep}{1.5pt}

    \begin{tabular}{llllccccccc}
        \toprule
        \multirow{2}{*}{(1)} & \multirow{2}{*}{(2)} & \multirow{2}{*}{(3)} & \multirow{2}{*}{(4)} & \multirow{2}{*}{(5)} & \multicolumn{3}{c}{Poses} &                  & \multicolumn{2}{c}{Depth}                                                             \\ \cline{6-8} \cline{10-11}
        \rule{0pt}{8pt}      &                      &                      &                      &                      & ATE $\downarrow$          & RTE $\downarrow$ & RRE $\downarrow$          &  & Abs Rel $\downarrow$ & $\delta\textless 1.25 \uparrow$ \\
        \midrule
        \checkmark           & \checkmark           & \checkmark           & \checkmark           & \checkmark           & 0.451                     & 0.172            & 1.173                     &  & 0.513                & 34.27                           \\
                             & \checkmark           & \checkmark           & \checkmark           & \checkmark           & 0.466                     & 0.150            & 1.267                     &  & 0.518                & 33.75                           \\
                             &                      & \checkmark           & \checkmark           & \checkmark           & 0.491                     & 0.197            & 1.367                     &  & 0.520                & 33.37                           \\
                             &                      &                      & \checkmark           & \checkmark           & 0.465                     & 0.184            & 1.427                     &  & 0.573                & 32.40                           \\
                             &                      &                      &                      & \checkmark           & 0.501                     & 0.198            & 1.389                     &  & 0.569                & 32.10                           \\
                             &                      &                      &                      &                      & 0.524                     & 0.194            & 1.677                     &  & 0.550                & 31.99                           \\
        \bottomrule
    \end{tabular}
    \caption{Ablation study on Sintel dataset.}
    \label{tab:ablation}

\end{table}

\paragraph{(1) Impact of Second Stage Training.} The variant in the second row omits the second stage training on longer sequences, resulting in a general performance degradation. This highlights the importance of long-range finetuning to handle extended sequences and mitigate cumulative drift.

\paragraph{(2) Effect of Relative Pose Loss.} In the third row, removing $\mathcal{L}_{relpose}$ impairs pose estimation, highlighting its crucial role in ensuring trajectory consistency.

\paragraph{(3) Contribution of TDM.} The variant in the fourth row removes the TDM,  resulting in the most severe drop in depth accuracy. It shows that motion information is essential for capturing high-frequency motion details.

\paragraph{(4) Influence of PSM.} As shown in the fifth row, removing PSM severely degrades performance across most metrics. This result underscores the PSM's critical role as a long-term geometric anchor against global drift.

\paragraph{(5) Importance of TCA.} The variant in the sixth row removes the TCA. It leads to significant drop in pose accuracy.  This demonstrates that providing the memories with a rich temporal context is vital for resolving motion ambiguities.

\section{Conclusion}

In this work, we present Mem4D, a novel online framework for dynamic scene reconstruction from monocular video. Mem4D resolves the Memory Demand Dilemma in existing memory-based methods by decoupling the modeling of low-frequency static geometry and high-frequency dynamic motion into two separate memories: the Persistent Structure Memory that ensures long-term geometric stability, and the Transient Dynamics Memory that captures fine-grained motion details.  Extensive experiments demonstrate that Mem4D achieves impressive performance across multiple tasks, validating the effectiveness of our method.

\paragraph{Limitations}
As a feed-forward online method, Mem4D can still be susceptible to drift accumulation over extremely long video sequences. Integrating a Bundle Adjustment module could further enhance long-term accuracy. Besides, our approach relies on supervised training, yet dense 4D ground-truth data is scarce and difficult to acquire, which limits scalability and generalization. Exploring self-supervised training strategies could help address this issue.

\bibliography{aaai2026}
\end{document}